\documentclass[journal,twoside,web]{ieeecolor}
\usepackage{generic}
\usepackage{cite}
\usepackage{amsmath,amssymb,amsfonts}
\usepackage{algorithmic}
\usepackage{graphicx}
\usepackage{algorithm,algorithmic}
\usepackage{hyperref}
\hypersetup{hidelinks=true}
\usepackage{textcomp}
\def\BibTeX{{\rm B\kern-.05em{\sc i\kern-.025em b}\kern-.08em
    T\kern-.1667em\lower.7ex\hbox{E}\kern-.125emX}}
\usepackage{bm}
\usepackage{mwe} 
\usepackage{lipsum}
\usepackage{array}
\usepackage{multirow}
\usepackage{bbding}

\usepackage{changes}
\usepackage[utf8]{inputenc}
\usepackage{xcolor}

\newcommand{\etal}{\textit{et al}.}
\newcommand{\ie}{\textit{i}.\textit{e}.}
\newcommand{\eg}{\textit{e}.\textit{g}.}

\begin{document}
\title{DualStreamFoveaNet: A Dual Stream Fusion Architecture with Anatomical Awareness for Robust Fovea Localization}

\author{Sifan Song, Jinfeng Wang, Zilong Wang, Hongxing Wang, Jionglong Su, Xiaowei Ding, Kang Dang
	\thanks{This work was supported by the Key Program Special Fund in XJTLU (KSF-A-22) and VoxelCloud, Inc. \textit{(Corresponding Authors: Jionglong Su; Xiaowei Ding; Kang Dang)}}
        \thanks{Sifan Song is with VoxelCloud, Inc., Shanghai, 201114, China (e-mail: dalesong6@gmail.com)}
	\thanks{Jinfeng Wang and Jionglong Su are with School of AI and Advanced Computing, XJTLU Entrepreneur College (Taicang), Xi'an Jiaotong-Liverpool University, Suzhou, 215123, China (e-mail: Jf.Jacob.Wong@gmail.com, Jionglong.Su@xjtlu.edu.cn)}
	\thanks{Zilong Wang is with VoxelCloud, Inc., Shanghai, 201114, China (e-mail: dddwzl3703@163.com)}
        \thanks{Hongxing Wang is with School of Big Data and Software Engineering, Chongqing University, Chongqing, 400044, China (e-mail: ihxwang@cqu.edu.cn)}
	\thanks{Xiaowei Ding is with Shanghai Jiao Tong University, Shanghai, 200240, China (e-mail: dingxiaowei@sjtu.edu.cn)}
	\thanks{Kang Dang is with School of AI and Advanced Computing, XJTLU Entrepreneur College (Taicang), Xi'an Jiaotong-Liverpool University, Suzhou, 215123, China and VoxelCloud, Inc., Shanghai, 201114, China (e-mail: Kang.Dang@xjtlu.edu.cn)}
}

\maketitle

\begin{abstract}
	Accurate fovea localization is essential for analyzing retinal diseases to prevent irreversible vision loss. While current deep learning-based methods outperform traditional ones, they still face challenges such as the lack of local anatomical landmarks around the fovea, the inability to robustly handle diseased retinal images, and the variations in image conditions. In this paper, we propose a novel transformer-based architecture called DualStreamFoveaNet (DSFN) for multi-cue fusion. This architecture explicitly incorporates long-range connections and global features using retina and vessel distributions for robust fovea localization. We introduce a spatial attention mechanism in the dual-stream encoder to extract and fuse self-learned anatomical information, focusing more on features distributed along blood vessels and significantly reducing computational costs by decreasing token numbers. Our extensive experiments show that the proposed architecture achieves state-of-the-art performance on two public datasets and one large-scale private dataset. Furthermore, we demonstrate that the DSFN is more robust on both normal and diseased retina images and has better generalization capacity in cross-dataset experiments.
\end{abstract}


\section{Introduction}
\label{sec:introduction}

\IEEEPARstart{T}{he} fovea, a crucial feature of the retina situated at the center of the macula, plays a significant role in producing sharp and detailed vision~\cite{weiter1984visual}. Identifying the precise location of the macula and fovea is critical for detecting several retinal diseases, including age-related macular degeneration and diabetic maculopathy~\cite{aquino2014establishing, deka2015detection, medhi2016effective}. The distance between the fovea and abnormalities such as hemorrhages and exudates directly influences the severity of vision loss~\cite{medhi2016effective}. Therefore, early detection of the fovea's location is essential for preventing irreversible damage to vision~\cite{deepak2011automatic, giancardo2012exudate, deka2015detection}.

\begin{figure*}
	\centering
	\includegraphics[width=1.95\columnwidth]{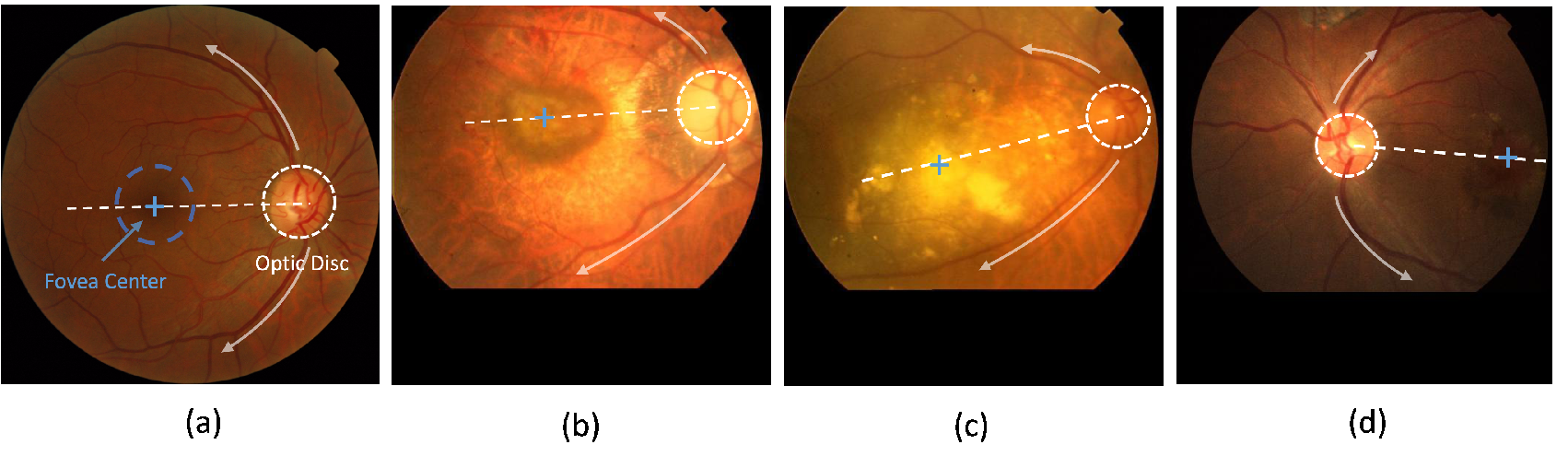}
	\caption{{Illustration of the shared global anatomical relationships among the fovea, optic disc, and main blood vessels, in normal (a),diseased (b, c), and poorly conditioned (d) retina images.}} \label{Fig_Moti}
\end{figure*}

Detecting the location of the fovea is crucial in preventing irreversible damage to vision and is an essential part of automated fundus diagnosis~\cite{deepak2011automatic, giancardo2012exudate, deka2015detection}. {Generally, the fovea is a small, central pit located in the macula region of the retina, responsible for sharp and detailed central vision. In retinal images, the fovea appears as a dark, circular area due to its high density of cone photoreceptors and lack of blood vessels, which enhances visual acuity. It is typically situated approximately 2.5 optic disc diameters temporal (toward the ear) from the center of the optic disc, aligned along the main axis of the retinal blood vessels (Fig.~\ref{Fig_Moti}-a)~\cite{asim2012detection, aquino2014establishing, deka2015detection}.} 

However, several challenges arise in localizing fovea accurately. First, the surrounding retinal tissue's color intensity makes the fovea's dark appearance indistinguishable, and there are no anatomical landmarks in its vicinity~\cite{sekhar2008automated, asim2012detection, medhi2016effective}. Second, retinal diseases such as hemorrhages and microaneurysms can obscure the distinction between the fovea and retinal background {(Fig.~\ref{Fig_Moti}-b and c)}, leading to inaccurate results~\cite{li2004automated, narasimha2006robust, geetharamani2018macula, guo2020robust}. Similarly, bright lesions like exudates can change the fovea's lightness to bright and lead to incorrect localization. Third, low light conditions and non-standard fovea locations during photography pose challenges in accurate fovea localization~\cite{pachade2019novel, xie2020end, song2022bilateral} {(Fig.~\ref{Fig_Moti}-d)}. Blurred and poorly lit images make estimating macula difficult, and images with centrally located optic discs instead of macula can lead to opposite-side predictions compared to the ground truth. Given these challenges, a robust fovea localization method must take into account global image features to deliver accurate result.

Fortunately, {although the image conditions or retina diseases may reduce the distinctness between fovea and macula regions}, {the global} anatomical structures outside the fovea, such as {optic disc and} blood vessels, have proved to be useful for fovea localization in various studies~\cite{li2004automated, narasimha2006robust, sekhar2008automated, asim2012detection, gegundez2013locating, giachetti2013use, aquino2014establishing, deka2015detection, dashtbozorg2016automatic, girard2016simultaneous, medhi2016effective, molina2017fast, guo2020robust, song2022bilateral} and other lesion detection research~\cite{zhang2019detection, huang2022rtnet}. {The fovea center is typically situated along an imaginary line that extends from the center of the optic disc and runs through the macula. This line often aligns with the main superior and inferior temporal blood vessels, which branch out from the optic disc. Experienced clinicians utilize this spatial relationship between the optic disc and main vessels to approximate the fovea position when facing the aforementioned challenging retina images (Fig.~\ref{Fig_Moti}-b, c and d)~\cite{li2004automated, aquino2014establishing, wibawa2023methods}.}

Previous works have employed carefully designed morphological methods to model the intricate anatomical relationships between the fovea and blood vessels~\cite{li2004automated, aquino2014establishing, deka2015detection, medhi2016effective, guo2020robust}. However, these traditional methods based on morphological models may falter when faced with images exhibiting rare fovea positions and color intensity variations. While more recent studies have harnessed the power of deep learning to enhance performance, they typically rely solely on fundus images as input~\cite{sedai2017multi, al2018multiscale, geetharamani2018macula, meyer2018pixel, pachade2019novel, huang2020efficient, xie2020end, bhatkalkar2021fundusposnet} and are implemented on datasets containing few challenging images. These approaches result in three main shortcomings: 1) inadequate exploitation of the anatomical structure outside the macula, as only fundus images are used as input; 2) typical convolution-based architectures lacking the incorporation of global features; and 3) sensitivity to challenging cases, such as rare fovea positions and severe lesions.

To {address} these challenges, we introduce a novel architecture, DualStreamFoveaNet (DSFN), {to adequately utilize the crucial global anatomical relationship (as shown in Fig.~\ref{Fig_Moti})} for robust fovea localization. {For many medical and biomedical research areas, the fusion of multi-branch features has proved to be a crucial technique for integrating information from different cues~\cite{prakash2021multi, guo2022context, yang2023lightingnet, wang2023stfuse, guo2023variational, li2023learning}.} Drawing inspiration from TransFuser~\cite{prakash2021multi}, our design features a dual-stream encoder that fuses multi-cue features and a decoder that generates result maps. By incorporating images from distinct cues (i.e., fundus and vessel distribution) as inputs for the encoder, we effectively utilize the anatomical structure outside the macula. In the encoder, we merge the multi-cue features of fundus and vessel images using transformer-based modules, dubbed the Bilateral Token Incorporation (BTI), which enables the modeling of global features and long-range connections for fovea localization, ensuring robust performance even in challenging images. Unlike TransFuser, which directly reduces and recovers token numbers by applying average pooling and bilinear interpolation methods—resulting in information loss—we employ TokenLearner and TokenFuser~\cite{ryoo2021tokenlearner} within the BTI module. The attention mechanism of TokenLearner extracts self-learning spatial information from both cues, allowing our design to effectively exploit structural features along the optic disc and vessel distribution. Furthermore, the attention mechanism reduces the number of tokens in the BTI module, significantly cutting down on computational effort.

\textbf{Contributions:}
\begin{itemize}
	\item We present a novel approach to fovea localization called DualStreamFoveaNet (DSFN), which utilizes a multi-cue fusion architecture. Unlike traditional convolution-based methods, our approach incorporates transformer-based structures to integrate long-range connections from various cues on a global level.
	\item We introduce the Bilateral Token Incorporation (BTI) with learnable tokens to enhance the efficiency of transformer-based fusion. This adaptive learning of tokens dramatically reduces the token number from 1024 to 64, while the spatial attention mechanism of the learnable tokens concentrates more on features along the vessel distribution, resulting in robust fovea localization.
	\item Our proposed DSFN achieves state-of-the-art performance on three datasets (\texttt{Messidor}, \texttt{PALM}, and \texttt{Tisu}) at a mere 25\% computational cost (62.11G FLOPs) compared to the best previous work~\cite{song2022bilateral} (249.89G FLOPs), demonstrating superior performance and generalization capability in challenging cases.
\end{itemize}

\section{Related Work}
\subsection{Anatomical Structure-based Methods} 
Previous studies have typically relied upon classical image processing methodologies for the estimation of foveal regions. The anatomical correlation between the approximate macular location, optic disc (OD), and blood vessels served as the basis for this reliance~\cite{li2004automated, narasimha2006robust, aquino2014establishing, deka2015detection, medhi2016effective, guo2020robust, fu2022fovea}. Specifically, the fovea center is positioned approximately 2.5 OD diameters from the OD's center, lying on the symmetrical axis of the main vessel branches passing through the OD. These two anatomical landmarks, the distance relationship between the OD and fovea as well as the major vessel orientation, have been widely adopted in previous work for fovea localization relying on image processing techniques~\cite{li2004automated, narasimha2006robust, aquino2014establishing, medhi2016effective, guo2020robust, fu2022fovea}.

Some studies only rely on the location of OD to identify the foveal region. Narasimha~\etal~\cite{narasimha2006robust} propose a two-step approach that incorporates the distance from the OD center and image intensity to update the region of interest (ROI) before localizing the fovea center. Sekhar~\etal~\cite{sekhar2008automated} employ spatial relations to select a sector-shaped candidate ROI. The sector's boundaries extend 30 degrees above and below the line through the image's center and OD. Subsequently, they use a threshold to filter intensity and estimate the foveal region. In color fundus images, blood vessels appear relatively darker in contrast to the OD. 

Some studies only utilize the extracted vessels skeleton to estimate the ROI containing the macula. Deka~\etal~\cite{deka2015detection} and Medhi~\etal~\cite{medhi2016effective} divide the image into several horizontal strips and select the ROI based on the absence of blood vessels in the vicinity of the macula. Thresholding is then utilized to detect the macula. Guo~\etal~\cite{guo2020robust} propose a morphological technique to fit the segmented skeleton of major vessels using a parabolic curve. The parabola's axis of symmetry is employed to localize the foveal region. 

The anatomical relationship between OD and vessels has been utilized in fovea localization. Asim~\etal~\cite{asim2012detection} estimate the ROI based on the pre-detected OD location and minimum intensity values, excluding the ROI near the vascular tree to enhance accuracy. Li~\etal~\cite{li2004automated}, Aquino~\etal~\cite{aquino2014establishing} and Fu~\etal~\cite{fu2022fovea} also use a parabolic fit of major vessels to detect the orientation of the macula and exploit the anatomical correlation, specifically the distance, between the OD and foveal center for approximate localization. Notably, Fu~\etal~\cite{fu2022fovea} employ a deep learning method (U-Net~\cite{ronneberger2015u}) rather than image processing techniques to detect the OD and vessels. Nevertheless, the methods only relying on anatomical features may exhibit poor performance when confronted with pathological images. Furthermore, they generally exhibit less competitive performance in comparison to recent deep learning-based approaches.

Efforts have been made to localize the macula without reliance on anatomical features. For example, GeethaRamani and Balasubramanian~\cite{geetharamani2018macula} propose a method to segment the macula using an unsupervised clustering algorithm. Pachade~\etal~\cite{pachade2019novel} directly select the square at the image center as ROI and apply an intensity filter for fovea localization. However, these approaches, which do not consider anatomical features, may fail in cases of varying illumination or when the macula is not situated in the standard location (\ie, the image center). 

\begin{figure*}
	\centering
	\includegraphics[width=1.95\columnwidth]{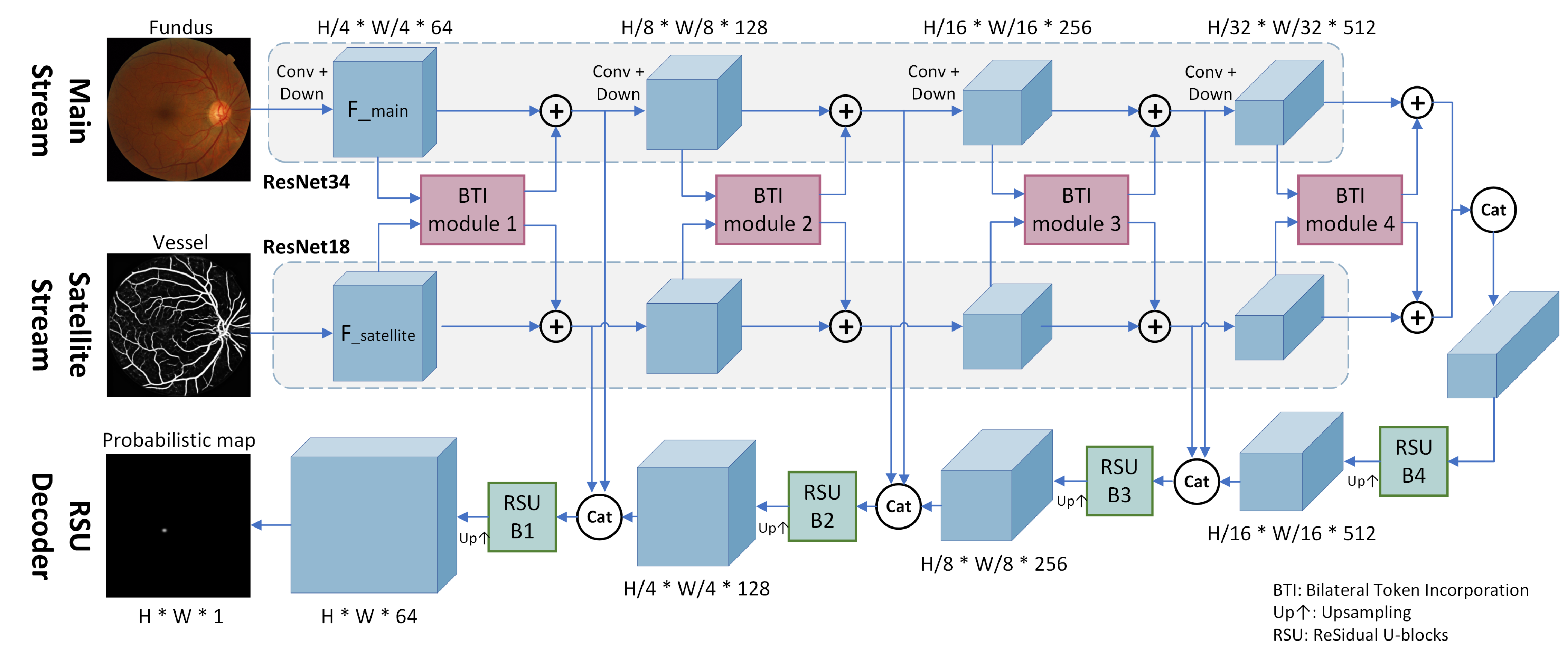}
	\caption{The overall architecture of our proposed DualStreamFoveaNet (DSFN) network. It consists of a two-stream encoder which effectively incorporates long-range features from both fundus and vessel distributions, and a decoder that generates accurate segmentation results by effective multi-scale feature incorporation.} \label{Fig_Arch}
\end{figure*}

\subsection{Deep Neural Networks in Fovea Localization} 
Deep learning has demonstrated superiority over traditional image processing and morphological techniques in many fields of medical image analysis, such as classification, segmentation and object localization~\cite{suzuki2017overview, zhang2020artificial, murtaza2020deep, qin2020u2, yu2021location, yang2021artificial}. Regarding the task of fovea localization, existing research can be broadly categorized into two primary domains: regression and segmentation. 

Many deep learning-based methods formulate fovea localization as a regression task. Al-Bander~\etal~\cite{al2018multiscale} and Huang~\etal~\cite{huang2020efficient} propose a two-step regression approach, initially predicting the ROI and subsequently inputting them into neural networks for fovea center localization. Meyer~\etal~\cite{meyer2018pixel} and Bhatkalkar~\etal~\cite{bhatkalkar2021fundusposnet} employ pixel-wise distance or heatmap regression approaches to jointly localize the OD and fovea. Xie~\etal~\cite{xie2020end} introduce a hierarchical regression network that integrates a self-attention mechanism in the process of fovea localization~\cite{cheng2016long, zhang2019self}. The network predicts the fovea center via a three-stage localization architecture that progressively crops features from coarse to fine.

In addition to regression, deep learning-based methods also leverage the image segmentation paradigm. Tan~\etal~\cite{tan2017segmentation} design a single 7-layer convolutional network to point-wise predict the foveal region from input image patches. Sedai~\etal~\cite{sedai2017multi} propose a two-stage image segmentation framework to segment the foveal region progressively. However, standard CNN-based architectures are limited by their fixed-size convolutional kernels, thereby failing to incorporate long-range features adequately. Consequently, these CNN-based architectures may fail under atypical light conditions and non-standard fovea positions, or in the absence of OD and vessel information due to lesions.

To address the constraint of limited receptive field, our prior work proposes a two-branch segmentation architecture (Bilateral-ViT~\cite{song2022bilateral}) to model long-range connections. This architecture utilizes a multi-head self-attention (MHSA) mechanism of transformer networks~\cite{vaswani2017attention, dosovitskiy2020image,chen2021transunet} in a novel way. In Bilateral-ViT, the main branch consisted of 12 consecutive MHSA layers constituting global retinal features, while an auxiliary vessel branch extracted multi-scale spatial information from vessel segmentation. The decoder simultaneously fused multi-cue features from fundus and vessels using multi-scale convolutions. While achieving state-of-the-art results on two publicly available datasets, \texttt{Messidor}~\cite{decenciere2014feedback} and \texttt{PALM}~\cite{55pk-8z03-19}, Bilateral-ViT suffers from lacking global multi-cue fusion and high computational costs. To overcome these limitations, we develop a new architecture called DualStreamFoveaNet (DSFN) incorporating a dual-stream encoder capable of globally fusing multi-cue features. Additionally, we introduce adaptively learnable tokens to effectively reduce computation demands.

\section{Methodology}
In this work, we introduce DualStreamFoveaNet (DSFN), a novel multi-cue fusion architecture for accurate and robust fovea localization (Fig.~\ref{Fig_Arch}). DSFN adopts a U-Net-like design with a dual-stream encoder comprising a main stream and satellite stream. Four intermediate Bilateral Token Incorporation (BTI) modules fuse global features from different cues. A decoder with Residual U-blocks (RSU)~\cite{qin2020u2} effectively combines features from both encoder streams. DSFN's transformer-based fusion of multi-cue features via efficient BTI modules maximizes anatomical guidance for accurate and efficient fovea localization.

\subsection{Overall Architecture} 
\label{subsec:arch}
The overall architecture of DSFN is depicted in Fig.~\ref{Fig_Arch}. The encoder consists of two streams: the main stream and the satellite stream, which use ResNet34 and ResNet18 backbones, respectively. {The main stream uses the larger ResNet34 to extract rich texture features from fundus images. In contrast, the satellite stream employs a simpler ResNet18 to extract key and sparse anatomical information from blood vessel distributions. Unlike the main stream that uses fundus images as input, the satellite stream takes a vessel segmentation map generated by a pre-trained model.} This pre-trained vessel segmentation model is built on the DRIVE dataset~\cite{staal2004ridge} using the TransUNet~\cite{chen2021transunet} architecture, which is identical to that used in~\cite{song2022bilateral}.

The dual-stream encoder, featuring four intermediate modules for multi-cue fusion, is inspired by PVT~\cite{wang2021pyramid} and TransFuser~\cite{prakash2021multi}. {As shown in Fig.~\ref{Fig_Arch}}, each stream's backbone is divided into four convolutional {stages}, comprising convolution, {batch normalization, ReLU,} and downsampling layers (Conv+Down). The resulting intermediate tensors ($\text{F}_{\mbox{\scriptsize{main}}}$ and $\text{F}_{\mbox{\scriptsize{satellite}}}$) are then fed into the BTI module, which includes a {modified} TokenLearner, $T$ consecutive Multi-Head Self-Attention (MHSA) layers, and a TokenFuser {(Fig.~\ref{Fig_TL})}. 

{Essentially, we have made four key improvements in the BTI module: 
1) We innovatively utilize it to integrate multi-cue features generated from the convolution neural network. After rethinking the spatial mechanism~\cite{park2018bam, woo2018cbam} used in TokenLearner, we find it well-suited for anatomical-based feature extraction. We adopt $8\times8$ learned tokens instead of the original 8 tokens in TokenLearner before each ViT block, as the BTI module needs to adaptively learn features significantly related to anatomical structures (e.g., vessel distribution) for fovea localization. The $8\times8$ tokens can retain sufficient spatial and other information for subsequent long-range dependency integration. 2) We progressively retain a greater number of channels (128, 256, 512, 1024) for the learned tokens in the deeper stages of the encoder's BTI module. This approach is based on the observation that as the CNN training progresses, more abstract and high-level information is extracted from each stream. Retaining a greater number of channels allows for better integration and preservation of this information, enabling a gradual and effective fusion process. 3) We concatenate the input fundus and vessel features from the two streams first and then use two consecutive $1\times1$ convolutional layers to remap the features along the channel. Since the non-vessel regions of concatenated vessel features are near zero, they explicitly provide clear clues for the spatial attention map to focus on anatomical structures. 4) After the TokenLearner, we innovatively use MHSA-based feature fusion to address the limitations of CNN-based feature fusion (\ie, limited receptive field and inductive bias of CNN). The MHSA block, through its self-attention mechanism, builds long-range dependencies, facilitating accurate fovea localization using global anatomical landmarks.}

{In the BTI module, anatomical-related features are adaptively learned, enabling efficient and effective fusion of multi-cue features from both fundus and vessel distributions by the subsequent multi-head self-attention mechanism. In other words, the adaptively learned features reduce the computational power required to encode long-range and global dependencies in the MHSA blocks, making it a practical choice for real-world applications.}. The output features are element-wise summed with skip-connected features and fed into the {next convolutional stage of each stream}. Additionally, these output features from the BTI module are forwarded to RSU blocks for decoding operations.

In contrast to the plain convolutional blocks commonly used in basic UNet decoders, the DSFN decoder employs four customized ReSidual U-blocks (RSU)~\cite{qin2020u2} for feature fusion. Each block is a U-shaped structure for effective multi-scale feature incorporation. Qin et al.~\cite{qin2020u2} demonstrate that the RSU block outperforms other embedded structures (e.g., plain convolution or residual-like, inception-like, and dense-like blocks) due to the enlarged receptive fields of the embedded U-shaped architecture. Furthermore, the superiority of the RSU structure as a decoder for incorporating multiple features has also been assessed by~\cite{song2022bilateral}. As illustrated in Fig.~\ref{Fig_Arch}, RSU B4 serves as the bottleneck block between the encoder and decoder. The input to the other three RSU blocks is a concatenation of three types of features: (i) multi-scale skip-connected features from the main stream, (ii) multi-scale skip-connected features from the satellite stream, and (iii) the hidden feature decoded by the previous RSU block.

\subsection{Bilateral Token Incorporation (BTI) modules}
Standard transformer-based architectures, such as Vision Transformer (ViT)~\cite{dosovitskiy2020image} and TransUNet~\cite{chen2021transunet}, typically divide the input image into 2D windows (\eg, a $16\times16$ grid) to generate tokens. These tokens are then fed into subsequent transformer layers to model long-range feature connectivity. However, the tokens are extracted individually from a fixed-size grid, which may result in uninformative or irrelevant features for visual understanding and increased computational expense. Recent architectures with multiple transformer stages~\cite{wang2021pyramid,prakash2021multi} have significantly more layers than the standard ViT, further exacerbating these issues.

\begin{figure}
	\centering
	\includegraphics[width=1.0\columnwidth]{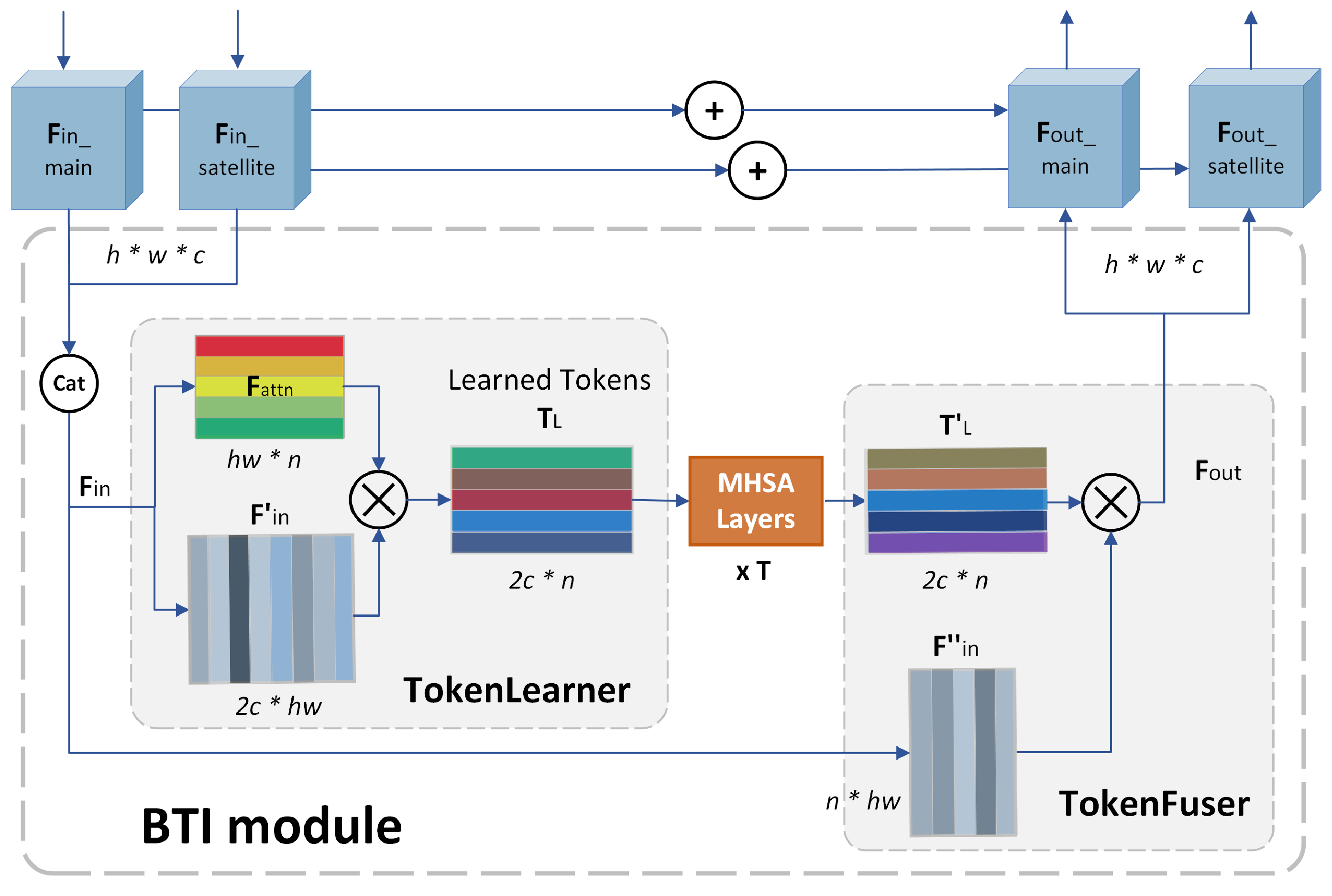}
	\caption{The structure of BTI module used in our DualStreamFoveaNet (DSFN). It contains TokenLearner, $t\times$ MHSA layers and TokenFuser. The $h$, $w$ and $c$ are height, width, and channel of the corresponding input features. The $n$ represents the number of learned tokens.} \label{Fig_TL}
\end{figure}

To address these challenges, the Bilateral Token Incorporation (BTI) module is introduced in the DSFN architecture (Fig.~\ref{Fig_TL}). The BTI module incorporates the TokenLearner~\cite{ryoo2021tokenlearner}, which adaptively learns tokens using a spatial attention mechanism. After being processed by Multi-Head Self-Attention (MHSA) layers (with head=8 and layer=12 for each BTI module), the tokens are remapped by the TokenFuser~\cite{ryoo2021tokenlearner} to the original input tensor dimensions (Fig.~\ref{Fig_TL}). Consequently, the BTI module offers two main advantages: (1) generating adaptive and learnable tokens to reduce the number of tokens, and (2) effectively merging and fusing features with long-range dependencies from both cues.

\subsubsection{TokenLearner} 
The TokenLearner modules leverage spatial attention to adaptively generate tokens from the input feature maps for the subsequent MHSA layers. Let $\text{F}^{\text{main}}_{\text{in}} \in \mathbb{R}^{h \times w \times c}$ and $\text{F}^{\text{satellite}}_{\text{in}} \in \mathbb{R}^{h \times w \times c}$ be the input tensors from the two streams, where $h$, $w$, and $c$ represent the height, width, and channel of the corresponding BTI module. As shown in Fig.~\ref{Fig_TL}, the concatenated feature $\text{F}_{\text{in}} \in \mathbb{R}^{h \times w \times 2c}$ {(Eq.~\ref{eq:conca})} is first fed into the TokenLearner. We customize the TokenLearner~\cite{ryoo2021tokenlearner} to contain a convolutional block containing two successive point-wise convolutional layers to reduce the dimensionality. This is followed by a flatten and a softmax operation to generate the spatial attention map $\text{F}_{\text{attn}} \in \mathbb{R}^{hw \times n}$ {(Eq.~\ref{eq:attention})}, where $n$ is the number of learned tokens. Moreover, $\text{F}_{\text{in}}$ is processed by a point-wise convolutional layer, which is then flattened and transposed to generate a transformed feature map $\text{F}'_{\text{in}} \in \mathbb{R}^{2c \times hw}$ {(Eq.~\ref{eq:prime})}. {Next, the learned tokens generated by spatial attention ($\text{F}_{\text{attn}}$) is given in Eq.~\ref{eq:lt}.}

\begin{equation}
	{\text{F}_{\text{in}} = \text{Concat}(\text{F}^{\text{main}}_{\text{in}}, \text{F}^{\text{satellite}}_{\text{in}})}
    \label{eq:conca}
\end{equation}
\begin{equation}
    \text{F}_{\text{attn}} = \text{Softmax}(\text{flatten}({g}(\text{F}_{\text{in}}))) \label{eq:attention}
\end{equation}
\begin{equation}
	\text{F}'_{\text{in}} = \text{Flatten}({h}(\text{F}_{\text{in}}))^T
    \label{eq:prime}
\end{equation}
\begin{equation}
	\text{T}_{\text{L}} = \text{F}'_{\text{in}} \text{F}_{\text{attn}}
    \label{eq:lt}
\end{equation}
where ${g}(\cdot)$ is a block with two point-wise convolutional layers, ${h}(\cdot)$ represents a point-wise convolutional layer, and the learned tokens are denoted as $\text{T}_{\text{L}} \in \mathbb{R}^{2c \times n}$. 

Through the spatial attention mechanism, the learned tokens represents an informative combination of features from corresponding spatial locations across the {concatenated} input cues. Given an input size of $512 \times 512$, in comparison to the 1024 tokens used in ViT and TransUNet~\cite{dosovitskiy2020image, chen2021transunet}, we only retain $8 \times 8$ tokens for each BTI module. Since the computation of MHSA is quadratic to the number of tokens, the computational cost is significantly decreased. As demonstrated in our experiments, TokenLearner enables us to not only significantly reduce the number of tokens but also extract features related to the distributions of blood vessel for accurate fovea localization. 

\subsubsection{TokenFuser}
The TokenFuser module reconstructs the tokens processed by the MHSA layers {(\ie, $\mathbf{T}'_{\text{L}}$ in Eq.~\ref{eq:mhsa})} back to their original resolution, allowing the output feature ($\text{F}_{\text{out}}$) to be further processed by the remaining stages of the DSFN architecture.  As shown in Fig.~\ref{Fig_TL}, the TokenFuser {fuses $\mathbf{T}'_{\text{L}}$} with the original input feature map ($\text{F}_{\text{in}}$) in the following way:

\begin{equation}
	{\mathbf{T}'_{\text{L}} = \text{MHSA}^{(t)} ( \text{MHSA}^{(t-1)} ( \cdots \text{MHSA}^{(1)}(\mathbf{T}_{\text{L}})\cdots)), t=12 }
    \label{eq:mhsa}
\end{equation}
\begin{equation}
	\text{F}''_{\text{in}} = \text{Sigmoid}(r(\text{flatten}(\text{F}_{\text{in}})^T)
        \label{eq:f2}
\end{equation}
\begin{equation}
	\text{F}_{\text{out}} = \text{Reshape}((\text{T}'_{\text{L}}\text{F}''_{\text{in}})^T)
        \label{eq:fout}
\end{equation}
\begin{equation}
	{\text{F}'^{\text{main}}_{\text{out}} = \text{F}_{\text{out}}[:, :, :c] \quad ; \quad \text{F}'^{\text{satellite}}_{\text{out}} = \text{F}_{\text{out}}[:, :, c:] }
 \label{eq:split}
\end{equation}
\begin{equation}
	{\text{F}^{\text{main}}_{\text{out}} = \text{F}'^{\text{main}}_{\text{out}} + \text{F}^{\text{main}}_{\text{in}} \quad ; \quad \text{F}^{\text{satellite}}_{\text{out}} = \text{F}'^{\text{satellite}}_{\text{out}} + \text{F}^{\text{satellite}}_{\text{in}} }
 \label{eq:res}
\end{equation}
where tensors $\text{F}_{\text{in}}, \text{F}_{\text{out}}\in\mathbb{R}^{h\times w\times 2c}$,  $\text{F}''_{\text{in}}\in\mathbb{R}^{n\times hw}$ and $\text{T}'_{\text{L}}\in\mathbb{R}^{2c\times n}$. ${r}(\cdot)$ represents a transformation involving two dense layers connected by an intermediate GeLU activation function. As we can see, the intermediate token representation $\text{T}'_{\text{L}}$ is remapped to $\text{F}_{\text{out}}$ which has the same resolution as the original input feature map $\text{F}_{\text{in}}$ {(Eq.~\ref{eq:f2} and Eq.~\ref{eq:fout})}. Next, $\text{F}_{\text{out}}$ is equally split to $\text{F}'^{\text{main}}_{\text{out}}$ {and} $\text{F}'^{\text{satellite}}_{\text{out}}$ {(Eq.~\ref{eq:split})}, representing outputs corresponding to the main and satellite streams, respectively. These reconstructed features are then element-wise summed with the skip-connected features from each stream (\ie, $\text{F}^{\text{main}}_{\text{in}}$ {and} $\text{F}^{\text{satellite}}_{\text{in}}$) to obtain the final output {of each stream ($\text{F}^{\text{main}}_{\text{out}}$ and $\text{F}^{\text{satellite}}_{\text{out}}$ in Eq.~\ref{eq:res})}.

\section{Experiments}
\subsection{Datasets and Network Configurations}
We first conduct experiments using the \texttt{Messidor}~\cite{decenciere2014feedback} and \texttt{PALM}~\cite{55pk-8z03-19} datasets. The \texttt{Messidor} dataset was developed for analyzing diabetic retinopathy and comprises 540 normal and 660 diseased retinal images. For this dataset, we utilize 1136 images fovea locations provided by~\cite{gegundez2013locating}. The \texttt{PALM} dataset was released for the Pathologic Myopia Challenge (PALM) 2019, which contains 400 images with fovea locations annotated. Of these, 213 images are pathologic myopia images, and the remaining 187 are normal retina images. For fairness of comparison, we follow the same data split as in the existing studies~\cite{xie2020end} and \cite{song2022bilateral}. 

Furthermore, we use a large-scale dataset (4103 images, named \texttt{Tisu}) which is a subset of data derived from a previously approved study, consists only of de-identified images to ensure subject privacy and data protection. Our current research accesses these images, maintaining confidentiality and ethical compliance. No identifiable subject information is included in the dataset used. Compared to the \texttt{Messidor} and \texttt{PALM} datasets, \texttt{Tisu} is more challenging as it contains a larger number of fundus images with various abnormalities besides hemorrhages, microaneurysms, and exudates. The ground-truth fovea centers are determined by averaging the labels provided by three medical experts. The dataset is split into training and testing sets in a 4:1 ratio.


One of our main contributions is the considerable reduction in FLOPs and GPU usage of our proposed DSFN. Specifically, compared to the previous state-of-the-art Bilateral-ViT~\cite{song2022bilateral}, our proposed DSFN consumes approximately 25\% FLOPs, 48\% GPU usage during training and 50\% GPU usage during inference. To assess whether the performance of Bilateral-ViT is influenced by its high computational requirements, we also introduce a light version, Bilateral-ViT/Lit (\ie, Bi-ViT/Lit in Table \ref{table-lit}). The basic architecture of Bilateral-ViT is retained, but we reduce the number of middle channels in each convolutional block by half and decrease the number of tokens from $32\times32$ to $8\times8$. In this case, Bilateral-ViT/Lit demonstrates comparable FLOPs and GPU usage to our DSFN, ensuring a fair evaluation.

\subsection{Implementation Details and Evaluation Metrics}
To preprocess the fundus images, we first remove the uninformative black background, and then pad and resize the cropped image region to $512\times512$. We simultaneously perform normalization and data augmentation on the input images of the main branch and the vessel branch. To train our DSFN model, we generate circular fovea segmentation masks {as the ground-truth} from the annotated fovea coordinates. During the inference process, we employ the sigmoid function on the decoder's output to acquire the segmentation map of the foveal region. The final fovea location coordinates are obtained by calculating the median values of all pixels in the map.

\begin{table}\footnotesize
	\caption{Configuration Comparison of Bilateral-ViT models and the proposed DSFN.}\label{table-lit}
	\centering
	\resizebox{1.0\columnwidth}{!}{
			\begin{tabular}{l |c |c |c |c |c   }
				\hline
				Methods & Tokens & Image Size & GFLOPs & GPU/T$^{\rm a}$ & GPU/I$^{\rm b}$  \\
				\hline
				{Bi-ViT}~\cite{song2022bilateral} & $32\times32$ & $512^2$ & 249.89 & 16873 & 5459  \\
				{Bi-ViT/Lit}~\cite{song2022bilateral} & $8\times8$ & $512^2$ & 83.05 & 8653 & 3093  \\
				DSFN (\textbf{Ours}) & $8\times8$ & $512^2$ & {62.11} &	{8083} & {2727} \\
				\hline
			\end{tabular}
		}
		\footnotesize{$^a$ The GPU usage when training (MiB)}\\
		\footnotesize{$^b$ The GPU usage when inferencing (MiB)}\\
\end{table}

All experiments are implemented in PyTorch and conducted on one NVIDIA GeForce RTX TITAN GPU. We use the Adam optimizer~\cite{kingma2014adam} with a batch size of 2. The loss function {for predicted segmentation map and ground-truth} is a summation of dice loss and binary cross-entropy. The experimental setup for the DSFN architectures {and the splitting of training and test sets on the \texttt{Messidor} and \texttt{PALM} datasets follow the identical approach} as previously reported in~\cite{song2022bilateral,xie2020end}. For our proposed DSFN, we set the initial learning rate to $1e^{-3}$, which is gradually decayed to $1e^{-9}$ using the CosineAnnealingLR strategy over 300 epochs on \texttt{Messidor} and \texttt{PALM}. For the \texttt{Tisu} dataset, we set the initial learning rate to $6e^{-5}$ for all related experiments. {To reduce overfitting, we adopt a data augmentation strategy to increase the diversity of training data (Details can be found in APPENDIX B). In addition, Xavier weight initialization is utilized for MLP weights in MHSA layers, and normal weight initialization is utilized for MLP biases in MHSA layers.}

{To evaluate the models' performance, we employ the $R$ rule as the assessment metric. The $R$ rule is a standard evaluation protocol for fovea localization and widely utilized in many fovea localization studies~\cite{gegundez2013locating, meyer2018pixel, xie2020end, song2022bilateral}. This metric shows the success rate of predictions, where each successful prediction is determined by ensuring that the Euclidean distance between the ground-truth and predicted fovea location coordinates does not exceed a predefined threshold (\ie, optic disc radius $R$).}{} To provide a comprehensive evaluation, we report the accuracy for different evaluation thresholds from $1/4 R$ to $2R$ (\eg, $2R$ indicating that the predefined threshold values are set to twice the radius of the optic disc $R$). {Furthermore, we employ the widely-used mean error (Err) of Euclidean distance between predicted and annotated fovea coordinates to assess the overall performance.}
	
\begin{table*}\tiny
		\caption{Comparison with existing studies using the \texttt{Messidor} and \texttt{PALM} datasets based on the $R$ rule. The best and second best results are highlighted in bold and italics respectively.}\label{table-SOTA}
		\centering
		\resizebox{1.58\columnwidth}{!}{
			\begin{tabular}{l c c c c c c c c }
				\hline
				\texttt{Messidor} &DL$^{\rm \mbox{a}}$&MF$^{\rm \mbox{b}}$ & 1/4 R (\%) & 1/2 R (\%) & 1R (\%) & 2R (\%) \\
				\hline
				Gegundez-Arias \etal (2013)~\cite{gegundez2013locating}   &\XSolidBrush &\checkmark & 76.32 & 93.84 & 98.24 & 99.30 \\
				Giachetti \etal (2013)~\cite{giachetti2013use}  	 &\XSolidBrush&\checkmark		& - 		& - 		& 99.10 		& - \\
				Aquino (2014)~\cite{aquino2014establishing} 	 &\XSolidBrush&\checkmark 		& 83.01 	& 91.28 	& 98.24 	& 99.56 \\
				Dashtbozorg \etal (2016)~\cite{dashtbozorg2016automatic}  	 &\XSolidBrush&\checkmark		& 66.50 	& 93.75	& 98.87	& 99.58 \\
				Girard \etal (2016)~\cite{girard2016simultaneous} 	 &\XSolidBrush&\checkmark 		& - 		& 94.00 	& 98.00 	& - \\
				Molina-Casado \etal (2017)~\cite{molina2017fast}	 &\XSolidBrush&\checkmark		& - 		& 96.08 	& 98.58 	& 99.50 \\
				Al-Bander \etal (2018)~\cite{al2018multiscale}   &\checkmark/ C &\XSolidBrush	 		& 66.80 	& 91.40 	& 96.60 	& 99.50 \\
				Meyer \etal (2018)~\cite{meyer2018pixel}	  &\checkmark/ M &\XSolidBrush 	& 94.01 	& 97.71 	& 99.74 	& - \\
				GeethaRamani \etal (2018)~\cite{geetharamani2018macula}  &\XSolidBrush&\XSolidBrush 		& 85.00 	& 94.08 	& 99.33	& - \\
				Pachade \etal (2019)~\cite{pachade2019novel} 	 &\XSolidBrush&\XSolidBrush 		& - 		& - 		& 98.66 	& - \\
				Huang \etal (2020)~\cite{huang2020efficient}   &\checkmark/ C &\XSolidBrush 		& 70.10 	& 89.20 	& 99.25 	& - \\ 
				Xie \etal (2020)~\cite{xie2020end}  &\checkmark/ C &\XSolidBrush 	& {98.15} 	& \textit{99.74} 	& \textit{99.82} 	& \textbf{100.00} \\
				Bhatkalkar \etal (2021)~\cite{bhatkalkar2021fundusposnet}  &\checkmark/ M &\XSolidBrush & 95.33 & \textit{99.74} & \textbf{100.00} & - \\
				Bi-ViT (2022)~\cite{song2022bilateral}  &\checkmark/ M &\checkmark & \textit{98.59} & \textbf{100.00} & \textbf{100.00} & \textbf{100.00} \\
				Bi-ViT/Lit (2022)~\cite{song2022bilateral} &\checkmark/ M &\checkmark & 98.50 & \textbf{100.00} & \textbf{100.00} & \textbf{100.00} \\
				{Wibawa \etal (2023)~\cite{wibawa2023methods}}   &\XSolidBrush &\checkmark 		& - 	& - 	& {98.24} 	& - \\ 
				{Sigut \etal (2024)~\cite{sigut2024fovea}}   &\XSolidBrush &\checkmark 		& {92.69} 	& {98.94} 	& {99.56} 	& - \\ 
				\textbf{DSFN ({Ours})}  &\checkmark/ M &\checkmark & \textbf{98.86} & \textbf{100.00} & \textbf{100.00} & \textbf{100.00} \\
				\hline
				\hline
				\texttt{PALM}  &DL$^{\rm \mbox{a}}$&MF$^{\rm \mbox{b}}$& 1/4 R (\%) & 1/2 R (\%) & 1R (\%) & 2R (\%) \\
				\hline
				Xie \etal (2020)~\cite{xie2020end} 	  &\checkmark/ C &\XSolidBrush&-  	& - 	& {94} &- \\
				Bi-ViT (2022)~\cite{song2022bilateral}  &\checkmark/ M &\checkmark & \textit{65} & \textit{83} & \textit{96} &\textbf{98} \\
				Bi-ViT/Lit (2022)~\cite{song2022bilateral} &\checkmark/ M &\checkmark& 55 & 80 & 94 & \textit{96} \\
				\textbf{DSFN ({Ours})}  &\checkmark/ M &\checkmark& \textbf{69} & \textbf{85} & \textbf{97} &\textbf{98} \\
				\hline
			\end{tabular}
		}
		
		\footnotesize{$^{\rm \mbox{a}}$ Whether the method is based on deep learning (DF). {The \textbf{C} and \textbf{M} in this column represent that the method uses \textbf{C}oordinate regression and result \textbf{M}ap generation, respectively.}}\\
		\footnotesize{$^{\rm \mbox{b}}$ Whether the method is based on multi-cue features (MF), \eg, fundus images, vessels or optical discs.}\\
\end{table*}

\begin{table*}
		\caption{Comparison of performance on normal and diseased retinal images using the \texttt{Messidor} and \texttt{PALM} datasets. The best and second best results are highlighted in bold and italics respectively. }\label{table-path}
		
		\centering
		\resizebox{1.88\columnwidth}{!}{
			\begin{tabular}{l c c c c c c c c c c c c c}
				\hline
				\texttt{Messidor} & MF$^{\rm \mbox{a}}$ & Err$_\downarrow$ & \multicolumn{2}{c}{1/4 R($\%$)} & \multicolumn{2}{c}{1/2 R($\%$)} & \multicolumn{2}{c}{1R($\%$)} & \multicolumn{2}{c}{2R($\%$)}\\
				& & & Normal & Diseased & Normal & Diseased & Normal & Diseased & Normal & Diseased \\
				\hline
				UNet (2015)~\cite{ronneberger2015u} &\XSolidBrush &12.39& 95.15 & 93.33 & 97.76&95.00&97.95&95.33&97.95&95.33 \\
				
				U2 Net (2020)~\cite{qin2020u2} &\XSolidBrush &7.31& 98.51 & 97.33 & \textit{99.63} & 99.50 & \textit{99.63} & 99.50 & \textit{99.63} &99.50 \\
				
				TransUNet (2021)~\cite{chen2021transunet} &\XSolidBrush &7.61& 98.32 & 97.67 &\textbf{100.00}&\textit{99.83}&\textbf{100.00}&\textit{99.83}&\textbf{100.00}&\textit{99.83} \\
				
				Bi-ViT (2022)~\cite{song2022bilateral} &\checkmark &\textit{6.81}& 98.51& \textbf{98.67} &\textbf{100.00}&\textbf{100.00}&\textbf{100.00}&\textbf{100.00}&\textbf{100.00}&\textbf{100.00}\\
				
				Bi-ViT/Lit (2022)~\cite{song2022bilateral} &\checkmark &\textbf{6.77}&\textit{98.69} &\textit{98.33} &\textbf{100.00}&\textbf{100.00}&\textbf{100.00}&\textbf{100.00}&\textbf{100.00}&\textbf{100.00} \\
				
				\textbf{DSFN ({Ours})} &\checkmark &\textbf{6.77}& \textbf{99.07} & \textbf{98.67} &\textbf{100.00}&\textbf{100.00}&\textbf{100.00}&\textbf{100.00}&\textbf{100.00}&\textbf{100.00} \\
				\hline
				\hline
				\texttt{PALM} & MF$^{\rm \mbox{a}}$ & Err$_\downarrow$ & \multicolumn{2}{c}{1/4 R($\%$)} & \multicolumn{2}{c}{1/2 R($\%$)} & \multicolumn{2}{c}{1R($\%$)} & \multicolumn{2}{c}{2R($\%$)}\\
				& & & Normal & Diseased & Normal & Diseased & Normal & Diseased & Normal & Diseased \\
				\hline
				UNet (2015)~\cite{ronneberger2015u} &\XSolidBrush& 149.30& 74.47 & 18.87 & 76.60 &41.51 &76.60 &64.15 &78.72 & 73.58  \\
				
				U2 Net (2020)~\cite{qin2020u2} &\XSolidBrush& 62.62& \textit{93.62} & 28.30& \textit{95.74}& 60.38& \textit{97.87}&84.91&\textit{97.87} & \textbf{98.11} \\
				
				TransUNet (2021)~\cite{chen2021transunet} &\XSolidBrush& 104.38& \textbf{95.74} & 18.87 & \textbf{97.87}& 43.40& \textit{97.87}& 75.47&\textit{97.87}& 84.91\\
				
				Bi-ViT (2022)~\cite{song2022bilateral} &\checkmark& \textit{53.70}& \textbf{95.74} & \textit{37.74}& \textbf{97.87}& \textit{69.81}& \textbf{100.00}& \textit{92.45}& \textbf{100.00}& \textit{96.23} \\
				
				Bi-ViT/Lit (2022)~\cite{song2022bilateral} &\checkmark&62.47& 87.23 & 26.42 & {93.62}& 67.92& \textit{97.87} &90.57 &\textit{97.87}&{94.34} \\
				
				\textbf{DSFN ({Ours})} &\checkmark& \textbf{48.72}& \textbf{95.74} & \textbf{45.28} &	\textbf{97.87}&\textbf{73.58} &\textbf{100.00}&\textbf{94.34}&\textbf{100.00}&\textit{96.23} \\
				\hline
			\end{tabular}
		}
		
		\footnotesize{$^{\rm \mbox{a}}$ Whether the method is based on multi-cue features (MF), \eg, fundus images, vessels or optical discs.}\\
\end{table*}

\section{Results}
\subsection{Comparison to State-of-the-Art}

In Table~\ref{table-SOTA}, we compare the performance of DSFN with existing methods on the public dataset, \texttt{Messidor} and \texttt{PALM}. Methods are classified based on whether they use deep learning techniques and whether they incorporate multi-cue features. We observe that most traditional morphological methods~\cite{gegundez2013locating, giachetti2013use, aquino2014establishing, dashtbozorg2016automatic, girard2016simultaneous, molina2017fast} rely on landmarks outside the macula, such as vessels or the optic disc. However, in most deep learning-based studies, including coordinate regression methods~\cite{al2018multiscale, huang2020efficient, xie2020end} ($\checkmark/ C$ in Table~\ref{table-SOTA}) and segmentation methods~\cite{meyer2018pixel, bhatkalkar2021fundusposnet} ($\checkmark/ M$), only fundus images are used, resulting in poor incorporation of anatomical relationships throughout the entire image, leading to failure in more challenging cases. Bilateral-ViT~\cite{song2022bilateral} is the only previous deep learning-based method that incorporates fundus and vessel features for fovea localization, which achieves better accuracy and robustness. 
	
The proposed DSFN, which combines a transformer-based multi-cue fusion encoder and adaptive learning tokens, outperforms all previous studies in terms of fovea localization accuracy on the \texttt{Messidor} and \texttt{PALM} datasets. Specifically, in Table~\ref{table-SOTA}, DSFN achieves the highest accuracy of 98.86\% at $1/4 R$, with gains of 0.71\% and 3.53\% compared to previous works~\cite{xie2020end} and \cite{bhatkalkar2021fundusposnet}, respectively. Our network also achieves better performance than Bilateral-ViT~\cite{song2022bilateral} and its light version. At evaluation thresholds of $1/2 R$, $1 R$, and $2 R$, DSFN achieves 100\% accuracy, indicating a localization error of at most $1/2 R$ (approximately 19 pixels for an input image size of $512\times512$). 
	
The \texttt{PALM} dataset is more challenging, with a smaller number of images and complex diseased patterns. Our proposed DSFN demonstrates superiority over all other methods on this dataset in Table~\ref{table-SOTA}, achieving accuracies of 69\% and 85\% at $1/4 R$ and $1/2 R$, respectively, which are 4\% and 2\% better than Bilateral-ViT. In addition, our method achieves a 14\% improvement ($1/4 R$) over Bilateral-ViT/Lit, and a 3\% improvement ($1R$) over both Bilateral-ViT/Lit and \cite{xie2020end}. Therefore, our DSFN achieves state-of-the-art performance on both \texttt{Messidor} and \texttt{PALM} datasets with high computational efficiency. 
	
\begin{figure}
		\centering
		\includegraphics[width=0.90\columnwidth]{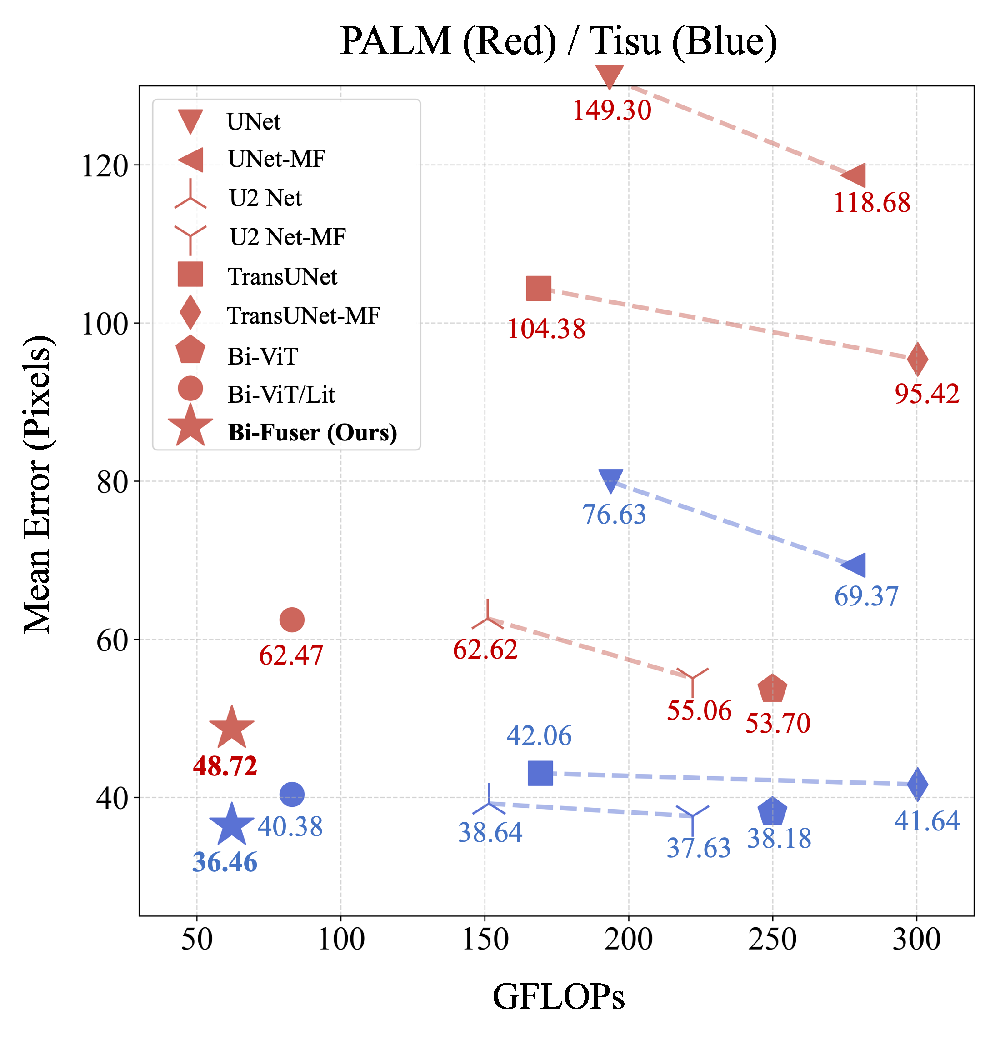}
		\caption{Visualization of mean errors ($Y$-axis) of different multi-cue fusion models. $X$-axis is the computational cost (GFLOPs). The red and blue markers are results on \texttt{PALM} and \texttt{Tisu} datasets, respectively. Numbers below the markers are corresponding mean errors.} \label{Fig-Mean}
	\end{figure}

	\begin{figure}
		\centering
		\includegraphics[width=0.95\columnwidth]{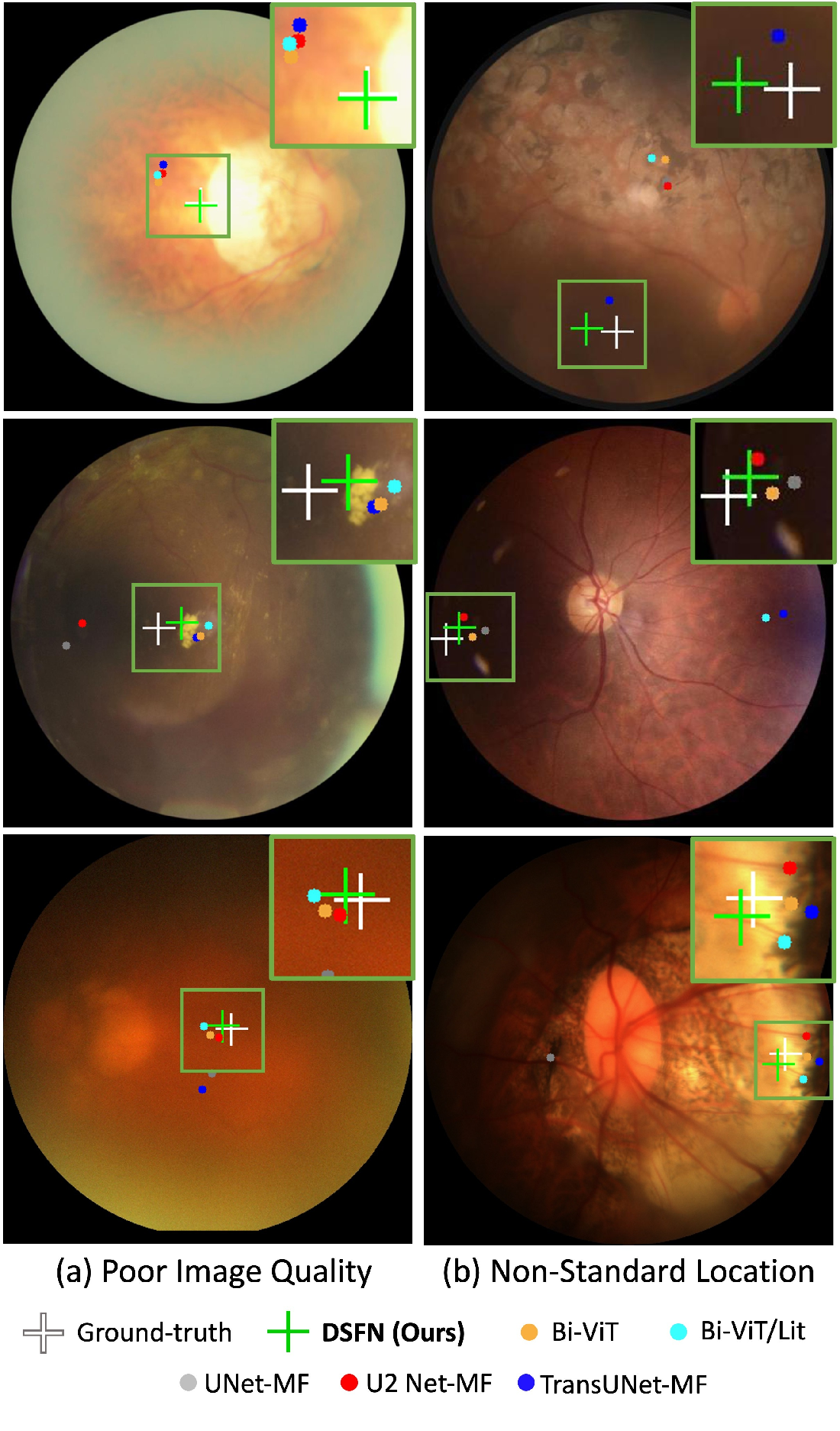}
		\caption{Comparing visual results of fovea localization predictions by various methods.} \label{Fig_1}
\end{figure}
	
\subsection{Fovea Localization on Normal and Diseased Images}
In Table~\ref{table-path}, we separately evaluate the performance of fovea localization for normal and diseased cases in \texttt{Messidor} and \texttt{PALM} datasets to assess the robustness of our method. We compare our proposed DSFN to several widely used segmentation networks, including UNet~\cite{ronneberger2015u}, U2 Net~\cite{qin2020u2}, and a hybrid version of TransUNet with ResNet50 for patch embedding~\cite{chen2021transunet}. The models of Bilateral-ViT, Bilateral-ViT/Lit and our DSFN are identical to those used in Table \ref{table-SOTA}. 
	
In Table~\ref{table-path}, the proposed DSFN achieves the lowest error (Err) on both datasets. Bilateral-ViT~\cite{song2022bilateral} and DSFN both obtain 100\% accuracy from $1/2 R$ to $2R$ on all the \texttt{Messidor} images, and achieve 100\% accuracy of $1R$ and $2R$ on normal \texttt{PALM} images. Compared to existing methods, DSFN demonstrates superior performance on almost all metrics on \texttt{Messidor}. As shown in Table~\ref{table-lit} and Table~\ref{table-path}, although DSFN has only 25\% FLOPs (62.11G) compared to Bilateral-ViT (249.89G), it achieves better performance on diseased images of \texttt{PALM} from $1/4 R$ to $2R$, with up to 7.54\% improvement compared to the other methods ($1/4 R$, Diseased). Its improvement is significantly increased to 18.86\% ($1/4 R$, Diseased) compared to Bilateral-ViT/Lit (83.05G, \ie, the network with the closest FLOPs to DSFN). Thus, our proposed DSFN is highly reliable in fovea localization on both normal and diseased fundus images with high computational efficacy. 
 
\subsection{Comparison of Multi-Cue Fusion Architectures}
To comprehensively assess the performance of models with input features from multiple cues (fundus and vessel distributions), we implement a multi-cue fusion version for the baseline models, UNet, U2 Net and TransUNet. We utilize two identical encoders, each with an input of a fundus image and a vessel map. The features are extracted independently and concatenated at the bottleneck for decoding (similar to \cite{sobh2018end, prakash2021multi}). These modified baseline models are referred to as UNet-MF, U2 Net-MF and TransUNet-MF. The results in Fig.~\ref{Fig-Mean} show that architectures using multi-cue features outperform typical networks with fundus-only input. This is evident in the challenging datasets, \texttt{PALM} and \texttt{Tisu}, where the improvement is pronounced. Moreover, the \texttt{Tisu} dataset is more complex than \texttt{PALM}, with more images (4103 \textit{vs.} 400) and a wider range of disease types and severity. Therefore, the results on \texttt{PALM} and \texttt{Tisu} demonstrate the potential of architectures that can effectively handle complex datasets.

\begin{table}\footnotesize
		\caption{{The performance of cross-dataset experiments. \textbf{Top}: The models trained on \texttt{Tisu} and tested on \texttt{PALM}. \textbf{Bottom:} The models trained on \texttt{PALM} and tested on \texttt{Tisu}. The models used here are exactly the corresponding ones used in Table~\ref{table-SOTA} and Fig.~\ref{Fig-Mean}. The best and second best results are highlighted in bold and italics respectively.}}\label{table-private}
		\centering
		\resizebox{1.0\columnwidth}{!}{
			\begin{tabular}{l c c c c c c c c c }
				\hline
				\texttt{Tisu}$\rightarrow$\texttt{PALM} & Err$_\downarrow$ & 1/4 R & 1/2 R & 1R & 2R (\%) \\
				\hline
				UNet-MF & 137.14 &52.25& 64.25&		77.25&	85.50 \\
				U2 Net-MF & 69.46 &55.00& \textbf{73.25}&	90.00&	97.25 \\
				TransUNet-MF & 78.98 &53.50& 71.75&	87.50&	95.75 \\
				Bi-ViT~\cite{song2022bilateral} & \textit{64.73} &\textbf{56.00}& \textit{72.75} & 90.00 & \textit{97.75} \\   
				Bi-ViT/Lit~\cite{song2022bilateral} & 71.14 & \textit{55.50}& \textbf{73.25} & \textit{90.75} & 97.00 \\   
				\textbf{DSFN ({Ours})} & \textbf{59.21} &55.25& \textbf{73.25} & \textbf{93.50} & \textbf{98.50} \\   
				\hline
				\hline
				{\texttt{PALM}$\rightarrow$\texttt{Tisu}} & Err$_\downarrow$ & 1/4 R & 1/2 R & 1R & 2R (\%) \\
				\hline
				UNet-MF~\cite{ronneberger2015u} & 155.96 & 19.01 & 27.74 & 36.51 & 44.55 \\
				U2 Net-MF~\cite{qin2020u2} & 129.86 & 18.82 & 30.83 & 47.96 & 64.10 \\
				TransUNet-MF~\cite{chen2021transunet} & 118.38 & 38.26 & 53.40 & 68.36 & 81.04 \\
				Bi-ViT~\cite{song2022bilateral} & \textit{86.92} & \textbf{40.92} & \textit{54.55} & \textit{70.97} & 83.65 \\  
				Bi-ViT/Lit~\cite{song2022bilateral} & {91.15} & 38.02 & 53.25 & 70.34 & \textit{85.40} \\   
				\textbf{DSFN ({Ours})} & \textbf{74.86} & \textit{39.34} & \textbf{55.38} & \textbf{75.68} & \textbf{89.42} \\   
				\hline
			\end{tabular}
		}
\end{table}

Fig.~\ref{Fig-Mean} shows a comparison of the described architectures (mean error against FLOPs) on \texttt{PLAM} (red markers) and \texttt{Tisu} (blue markers). Below each marker, we provide the corresponding mean error, and dashed lines connect each standard baseline model with its multi-cue fusion architecture (MF). The multi-cue fusion versions (UNet-MF, U2 Net-MF, and TransUNet-MF) outperform their standard versions at a considerably higher computational cost due to the additional encoder. For all multi-cue fusion architectures, the proposed  DSFN achieves the best results with the smallest errors on both \texttt{PLAM} (48.72 pixels) and \texttt{Tisu} (36.46 pixels). Furthermore, it demands only 62.11G FLOPs, which is four times less than the FLOPs required by the next best-performing approach, Bilateral-ViT (249.89G). Compared to the model with comparable FLOPs (Bilateral-ViT/Lit, 83.05G), DSFN shows significant advantages of 13.75 and 3.92 on \texttt{PLAM} and \texttt{Tisu}, respectively. 
	
Fig.~\ref{Fig_1} provides visual results of fovea localization on images with severe diseases from the \texttt{PALM} and \texttt{Tisu} datasets. These images in Fig.~\ref{Fig_1}-a and Fig.~\ref{Fig_1}-b suffer from poor image quality and non-standard fovea locations, respectively. Our DSFN generates the most accurate predictions for several challenging cases with poor lighting conditions and blurred appearance (Fig.~\ref{Fig_1}-a). For another challenging types (Fig.~\ref{Fig_1}-b), where the macula is close to the image boundary, the predictions of DSFN (green crosses) are closest to the ground-truth (white crosses). On the contrary, predictions from alternative architectures lacking the ability to globally incorporate long-range multi-cue features may lead to fovea locations appearing on the incorrect side of the optic disc (Fig.~\ref{Fig_1}-b). These results suggest that DSFN can adequately model fundus and vessel features of two streams, resulting in superior performance compared to other networks.

\begin{table}\footnotesize
		\caption{Comparison of performance between different inputs for the main and satellite streams on \texttt{PALM} and \texttt{Tisu}. The best and second best results are highlighted in bold and italics.}\label{table-inputs}
		\centering
		\resizebox{1.00\columnwidth}{!}{
			\begin{tabular}{l c c c c c c c c }
				\hline
				\texttt{PALM} & Err$_\downarrow$ & 1/4 R & 1/2 R & 1R & 2R (\%) \\
				\hline
				\textbf{Fundus+Vessel ({Ours})} & \textbf{48.72} & \textbf{69} & \textbf{85} & \textbf{97} & \textbf{98} \\  
				Fundus-only & {54.46} & {64} & \textbf{85} & \textit{95} & \textbf{98} \\   
				Vessel-only & 72.25 & 57 & {75} & 92 & \textit{97} \\   
				{Fundus+Vessel$^*$} & \textit{49.61} & \textit{67} & \textit{84} & \textit{95} & \textbf{98} \\
				\hline
				\hline
				\texttt{Tisu} & Err$_\downarrow$ & 1/4 R & 1/2 R & 1R & 2R (\%) \\
				\hline
				\textbf{Fundus+Vessel ({Ours})} & \textbf{36.46} & \textbf{52.32} & \textbf{75.49} & \textbf{92.93} & \textbf{97.44} \\
				Fundus-only & {37.48} & \textbf{52.32} & \textit{73.78} & {91.10} & \textit{96.95} \\   
				Vessel-only & 48.13 & {33.66} & 61.83&	89.39&	96.83\\   
				{Fundus+Vessel$^*$} & \textit{37.34} & \textit{51.59} & 73.66 & \textit{91.95} & \textbf{97.44} \\
				\hline
			\end{tabular}
		}
		
		{\footnotesize{$*$} denotes that the input of the satellite stream is pre-trained and \\ generated by UNet (\ie, a weaker cue).}\\
\end{table}
	
\subsection{{Performance of Cross-Dataset Experiments}}
In our cross-dataset experiments evaluating the generalization capabilities of the proposed DSFN, we focus on two demanding datasets—\texttt{PALM} and \texttt{Tisu}, characterized by a higher prevalence of retinal diseases and lower image quality. These datasets were interchangeably assigned as training and test sets. Notably, in Table \ref{table-private}-\textbf{Top}, DSFN exhibits substantial improvements, achieving reductions of 5.52 and 11.93 pixels in average localization error at the original image resolution compared to the previously top-performing method, Bilateral-ViT, and its lighter version, respectively. Additionally, in Table \ref{table-private}-\textbf{Bottom}, where models were trained on a smaller dataset (\texttt{PALM}) and tested on a larger, more challenging one (\texttt{Tisu}), DSFN demonstrates significant enhancements of 12.06 and 16.29 pixels in average location error compared to Bilateral-ViT and its lighter version, respectively. These findings underscore the DSFN's adaptability to diverse and previously unseen cases, mitigating overfitting concerns and highlighting its robust generalization capability.
	
\subsection{Ablation Study}
\subsubsection{Comparison of Inputs for DSFN}
Table \ref{table-inputs} compares the performance of DSFN when using different inputs. The standard input configuration uses fundus images and vessel maps as inputs for the main and satellite streams, respectively (Fundus+Vessel). These experiments achieve the best accuracy on all metrics, with the smallest mean error on both \texttt{PALM} (48.72 pixels) and \texttt{Tisu} (36.46 pixels). When using fundus images as the second input (Fundus+Fundus), the model's performance slightly degrades as feeding fundus images into the satellite stream does not provide the explicit anatomical structure for the BTI module to learn where to focus its attention. Experiments using vessel maps as both inputs (Vessel+Vessel) lead to significant accuracy decreases on both \texttt{PALM} and \texttt{Tisu} by 23.53 and 11.67 pixels, respectively, indicating a severe loss of information. In addition, we conduct experiments where the input of the satellite stream (\ie, vessel distribution map) is pre-trained and generated using a weaker architecture, UNet (Fundus+Vessel$^*$), as opposed to TransUNet in the default setting (Fundus+Vessel). These experiments on \texttt{PALM} and \texttt{Tisu} demonstrate the performance of our DSFN when the vessel input serves as a weak cue. The outcomes presented in Table~\ref{table-inputs} suggest that the performance of Fundus+Vessel$^*$ generally falls between that of Fundus+Vessel and Fundus-only. This finding demonstrates that even when providing a weak cue, the DSFN effectively learns valuable anatomical-related features, instead of starting to learn spatial attention from scratch (\ie, Fundus-only).

\begin{figure}
		\centering
		\includegraphics[width=0.98\columnwidth]{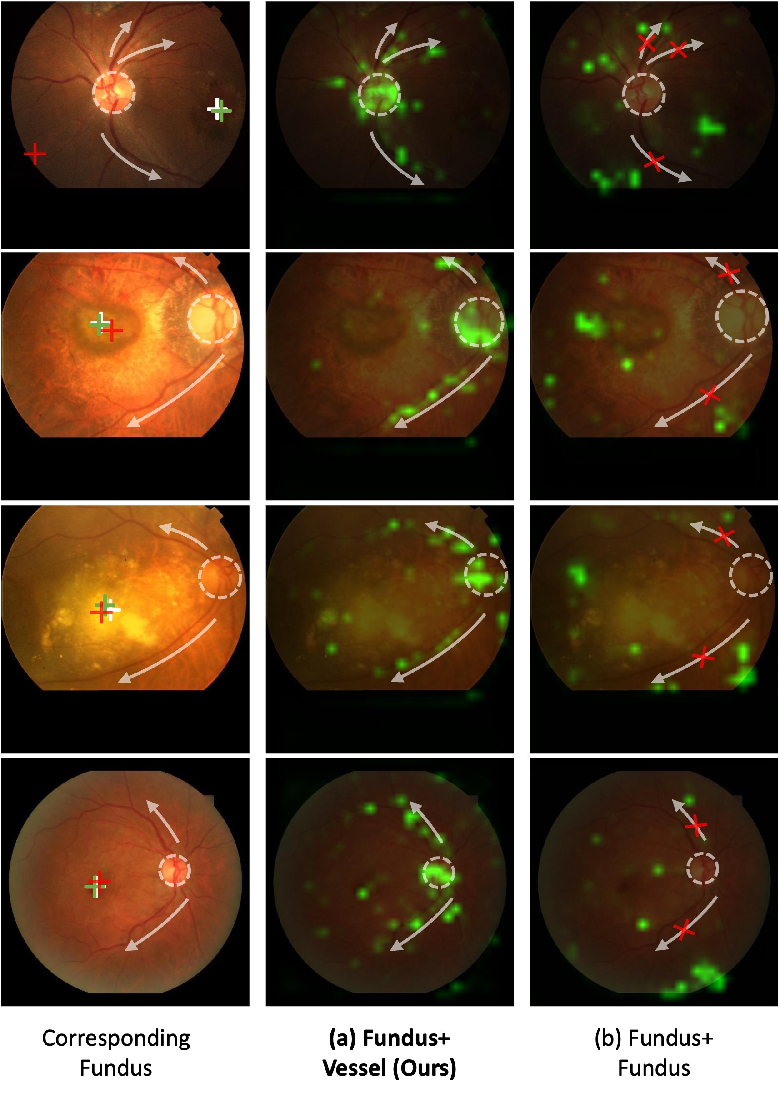}
		\caption{Visualizations generated from spatial attention maps, indicating the focus of TokenLearner in BTI module. These visual results have been resized and superimposed onto the the corresponding fundus. Circles with dotted lines represent the position of optic discs, and arrows represents the direction of main vessel branches. {The white, green, and red crosses in the first column are the ground-truth, prediction result with the second input of vessel (a) and fundus (b), respectively.}} \label{Fig-SA2}
\end{figure}
 
We utilize spatial attention maps to elucidate the self-learning features extracted by TokenLearners, visually demonstrating these weight maps in Fig.~\ref{Fig-SA2}. To visualize the attention of tokens in an element-wise manner, we maximize the saliency values among all the tokens $n$ for each position ${h\times w}$ of the spatial attention map $\text{F}_{\text{attn}} \in \mathbb{R}^{h \times w \times n}$ (refer to Eq.~\ref{eq:attention}). Subsequently, we normalize these values and superimpose the resulting maps onto the corresponding fundus, facilitating a comparison of their structural relationships.
	
In our experiments using Fundus+Vessel inputs, the spatial attention maps focus on structural features along the optic disc and the direction of vessel branches (Fig.~\ref{Fig-SA2}-a). This focus is feasible {and consistent} to the significant anatomical relationships between these structures and the fovea region {(shown in Fig.~\ref{Fig_Moti})}~\cite{li2004automated,aquino2014establishing,narasimha2006robust,sekhar2008automated, asim2012detection}. In contrast, despite having more detailed information in Fundus+Fundus experiments, TokenLearners, which are not guided by explicit anatomical structures, fail to learn features along the vessels (Fig.~\ref{Fig-SA2}-b). This results in fewer tokens carrying useful features for fovea localization and may limit the effectiveness of the intermediate BTI modules in DSFN, leading to a slight underperformance on \texttt{PALM} and \texttt{Tisu} datasets (Table \ref{table-inputs}). {In the first column of Fig.~\ref{Fig-SA2}, the predictions of Fundus+Vessel (green crosses) show superior results compared to those of Fundus+Fundus (red crosses).} Therefore, the structural information provided by vessel maps as the second input is crucial for achieving accurate fovea localization.
	
\subsubsection{Comparison of Methods for Reducing and Recovering Tokens}

\begin{figure}
		\centering
		\includegraphics[width=0.835\columnwidth]{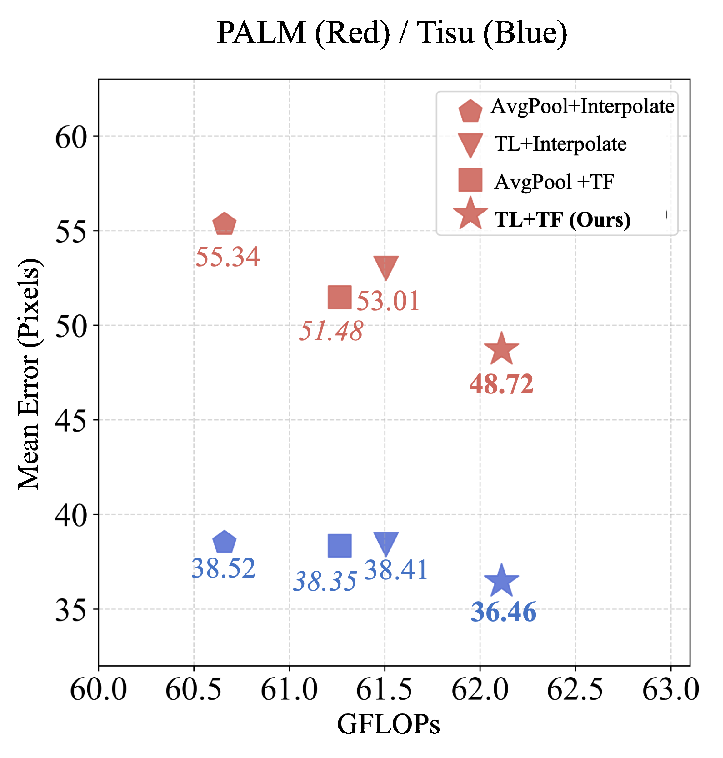}
		\caption{Visualization of mean errors ($Y$-axis) of ablation studies for reducing and recovering tokens. $X$-axis is the computational cost (GFLOPs). The red and blue markers are results on \texttt{PALM} and \texttt{Tisu} datasets, respectively. Numbers below the markers are corresponding mean errors.} \label{Fig-Mean2}
\end{figure}

\begin{table}\footnotesize
		\caption{{Comparison of performance between different configurations of learned tokens and layer number of MHSA blocks utilized in the BTI module. The best results are highlighted in bold.}}\label{table-tokens}
		\centering
		\resizebox{1.00\columnwidth}{!}{
			\begin{tabular}{c c c c c c c c c c c }
				\hline
				Optimal & Tokens & Layers & Err$_\downarrow$ & 1/4 R & 1/2 R & 1R & 2R (\%) \\
				\hline
				\XSolidBrush & 8 & 12 & 54.40 & 60 & 81 & 93 & 97 \\  
				\XSolidBrush & $4\times4$ & 12 & 52.45 & 61 & 78 & 93 & 97 \\  
				\checkmark & $8\times8$ & 12 & \textbf{48.72} & \textbf{69} & \textbf{85} & \textbf{97} & \textbf{98} \\  
				\XSolidBrush & $16\times16$ & 12 & 52.66 & 64 & {84} & 94 & \textbf{98} \\  
				\hline
				\hline
				\XSolidBrush & $8\times8$ & 8 & 49.96 & 66 & 82 & {96} & \textbf{98} \\  
				\XSolidBrush & $8\times8$ & 16 & 55.06 & {68} & \textbf{85} & 95 & 97 \\  
				\hline
			\end{tabular}
		}
\end{table}
 
To evaluate the effectiveness of different components within our BTI module for reducing and recovering token numbers, we perform a comprehensive set of ablation experiments on the \texttt{PALM} and \texttt{Tisu} datasets. Instead of employing TokenLearner (TL) and TokenFuser (TF) with adaptively learnable parameters, we alternatively test more straightforward methods used in~\cite{prakash2021multi}, average pooling (AvgPool) and bilinear interpolation (Interpolate), respectively.
	
In Fig.~\ref{Fig-Mean2}, we demonstrate a visualization of the mean error against the computational cost (FLOPs) on \texttt{PALM} (red) and \texttt{Tisu} (blue) datasets. Experiments using both TokenLearner and TokenFuser (TL+TF) achieve the best performance on \texttt{PALM} and \texttt{Tisu} with mean errors of 48.72 and 36.46 pixels, respectively. This is in contrast to using more straightforward methods such as average pooling (AvgPool) and bilinear interpolation (Interpolate), which may lead to a loss of information and reduced performance. In the proposed DSFN (TL+TF), total excess costs of FLOPs for the four BTI modules utilizing TokenLearner and TokenFuser are only \textbf{0.85G} and \textbf{0.61G}, respectively. This slight increase in computation leverages significant performance benefits, demonstrating the high efficacy of the adaptively learnable parameters of TokenLearner and TokenFuser in our architecture.

\subsubsection{{Comparison of Hyperparameter Settings}}
{The above subsections have evaluated the necessity of vessel distribution inputs and of the BTI module. We also conduct ablation studies to evaluate, in the BTI module, how the performance of feature integration and fusion is affected by the structure of the MHSA blocks, to ensure optimal configurations.}

{Table \ref{table-tokens} presents a comparative analysis of the performance on \texttt{PALM} between different configurations of the number of learned tokens and Multi-Head Self-Attention (MHSA) layers utilized in the BTI module. The table evaluates various configurations based on several metrics: mean error, and success rates at different thresholds (1/4 R, 1/2 R, 1R, and 2R). The optimal configuration identified is the use of $8 \times 8$ learned tokens with 12 MHSA layers, which achieves the best results across all metrics, including the lowest mean error (48.72) and the highest success rates at all thresholds. Other configurations, such as $4 \times 4$ tokens with 12 layers and $8 \times 8$ tokens with 8 layers, show lower performance, indicating that both the number of tokens and the depth of MHSA layers significantly influence the performance of feature integration and the accuracy of fovea localization.}

\section{Conclusions}
{\textbf{Summary}}. Accurate detection of the fovea is crucial for diagnosing retinal diseases. While anatomical structures outside the fovea, such as blood vessel distribution, are related to the fovea, few deep learning approaches exploit them to enhance fovea localization performance. In this paper, we propose a novel architecture, DualStreamFoveaNet (DSFN), which effectively fuses features from the retina and corresponding vessel distribution to achieve robust fovea localization. The DSFN comprises a two-stream encoder for multi-cue fusion and a decoder for generating fovea localization maps. Specifically,  We introduce the Bilateral Token Incorporation (BTI) with learnable tokens to enhance the efficiency of transformer-based fusion from both fundus and vessel images. Comprehensive experiments demonstrate the advantages of using DSFN, including more accurate localization results {through enhanced Multi-Cue Fusion}, insensitivity to diseased images, and computational {efficiency}. Furthermore, our proposed architecture sets a new state-of-the-art on two public datasets (\texttt{Messidor} and \texttt{PALM}) and one large-scale private dataset (\texttt{Tisu}) with metrics ranging from 1/4 $R$ to $2R$. Additionally, it outperforms other methods in cross-dataset experiments, showcasing superior generalization capacity {and robustness across datasets}.

{\textbf{Limitations and Future Works}. Although the promising performance of the proposed DSFN shown in our extensive experiments, the quality of the pseudo-mask of vessel structure still slightly affects the final fovea localization performance, as shown in Table~\ref{table-inputs}. Compared to a fuzzy fovea region with an indistinguishable appearance from nearby textures, vessels are very distinct structures with high contrast. In this paper, the significant performance increase of DSFN relies on the vessels' pseudo-masks generated by the TransUNet, which was trained on a small-scale vessel dataset. Recent studies~\cite{jiang2024covi, ding2024rcar, tan2024deep} have specifically focused on vessel structure prediction, potentially providing more accurate pseudo-masks of vessels for our architecture in future research. Moreover, recently emerged Segment Anything Model (SAM)-based segmentation foundation models~\cite{kirillov2023segment, ke2024segment}, and their medical fine-tuned versions~\cite{cheng2023sam, ma2024segment}, have shown significant disadvantages in predicting regions with no distinct boundaries and textures. Therefore, our design (\ie, utilizing global vessel structures to explicitly assist fovea localization) remains promising in this area.}

{Furthermore, the design of DSFN can be utilized as a general architecture for global feature integration and anatomical structure-guided segmentation tasks, as the input of the main and second streams can be any related images. In future work, we may upgrade DSFN for more medical imaging areas. In clinical practice, there are other tasks where even experienced clinicians require anatomical structure-based disease diagnosis. For example, in lung cancer screening, CT scans are commonly used for early detection. With the help of rib structures, doctors can better locate and observe lung lesions. In pelvic MRI, the structure of the pelvis can assist in the detailed evaluation of nearby pelvic organs and aid in diagnosing endometriosis and pelvic tumors.}

\section*{Appendix}
\subsection{Notation Summary} 
{All notions and abbreviations of this paper have been summarized in Table~\ref{table-abv}.}

\subsection{Augmentation Settings} 

{For all input images, to reduce overfitting, we adopt a data augmentation strategy utilizing a PyTorch package, albumentations~\cite{2018arXiv180906839B}. The details are shown as follows:}

{RandomResizedCrop (height = 512, width = 512, scale = (0.95, 1.05), ratio = (0.95, 1.05), p = 0.25), HorizontalFlip (p = 0.25), VerticalFlip (p = 0.25), ShiftScaleRotate (shift\_limit = 0.0625, scale\_limit = 0.05, rotate\_limit = 45, p = 0.25), OneOf ([Blur (blur\_limit = 5), GaussianBlur (blur\_limit = 5), MedianBlur (blur\_limit = 5), MotionBlur (blur\_limit = 5)], p = 0.25), RandomBrightnessContrast (brightness\_limit = 0.1, contrast\_limit = 0.1, p = 0.25), OneOf ([CLAHE (clip\_limit = 2), IAASharpen(), JpegCompression(), GaussNoise()], p = 0.25), CoarseDropout (p = 0.1).}

\begin{table*}\footnotesize
\caption{{Notation Summary}}\label{table-abv}
\centering
\resizebox{2.0\columnwidth}{!}{
\begin{tabular}{l | l  }
\hline
Abbreviation & Explanation  \\
\hline
{Bi-ViT} & Bilateral-ViT: an existing network~\cite{song2022bilateral} for fovea localization with bilateral feature fusion design \\
{Bi-ViT/Lit} & The light version of Bilateral-ViT \\
BTI & Bilateral Token Incorporation: The modules designed in this paper for globally bilateral feature fusion with high efficacy \\
CNN & Convolutional Neural Networks \\
DSFN & DualStreamFoveaNet: The two-stream architecture proposed in this paper \\
(G)FLOPs & (Giga) Floating Point Operations \\
MF & Multi-cue Features \\
MHSA & Multi-Head Self-Attention \\
OD & Optic Disc \\
ROI & Region of Interest \\
RSU & ReSidual U-blocks: The U-shape-based blocks utilized in the decoder of DSFN to integrate received features \\
TL and TF & TokenLearner and TokenFuser \\
ViT & Vision Transformer \\
\hline
\hline
Notation & Description  \\
\hline
$\text{F}^{\text{main}}_{\text{in}}, \text{F}^{\text{satellite}}_{\text{in}}$ & Input tensors from the main and satellite streams, respectively \\
$\text{F}_{\text{in}}$ & Concatenated input feature \\
$\text{F}_{\text{attn}}$ & Spatial attention map \\
$\text{F}'_{\text{in}}$ & Remapped input feature to generate $\text{T}_{\text{L}}$ \\
$\text{T}_{\text{L}}$ & Learned tokens \\
$\text{T}'_{\text{L}}$ & Intermediate token representation after MHSA layers \\
$\text{F}''_{\text{in}}$ & Remapped input feature to generate $\text{F}_{\text{out}}$ \\
$\text{F}_{\text{out}}$ & Output features generated by $\text{T}'_{\text{L}}$ and $\text{F}''_{\text{in}}$\\
$\text{F}'^{\text{main}}_{\text{out}}, \text{F}'^{\text{satellite}}_{\text{out}}$ & Output features equally split by $\text{F}_{\text{out}}$ \\
$\text{F}^{\text{main}}_{\text{out}}, \text{F}^{\text{satellite}}_{\text{out}}$ & Final output features for the main and satellite streams, respectively \\
\hline
\end{tabular}
}
\end{table*}

\section*{References}

\bibliographystyle{IEEEtran}
\bibliography{main}

\end{document}